\pdfoutput=1

\documentclass[11pt]{article}
\usepackage{graphicx}
\usepackage{subcaption}
\usepackage{xcolor}
\usepackage{amsmath}
\usepackage{listings}
\usepackage{amssymb}
\usepackage{soul}
\usepackage[]{acl2024}
\usepackage{times}
\usepackage{latexsym}
\usepackage{graphicx}
\usepackage{enumitem}
\usepackage{xcolor}
\usepackage[framemethod=TikZ]{mdframed}
\newmdenv[%
    backgroundcolor=gray!10,
    linecolor=black,
    outerlinewidth=0.5pt,
    roundcorner=1mm,
    skipabove=\topsep,
    skipbelow=\topsep,
    font=\ttfamily\small,
]{promptbox}
\usepackage[T1]{fontenc}

\usepackage[utf8]{inputenc}

\usepackage{microtype}

\usepackage{inconsolata}
\usepackage{xspace}
\newcommand{\metriclong}{Normalized Invariability to Choice of Examples}
\newcommand{\metric}{\text{NICE}\xspace}

\title{\metric: To Optimize In-Context Examples or Not?}

\author{Pragya Srivastava\thanks{\hspace{0.15cm}These authors contributed equally.},\hspace{0.15cm} {\bf Satvik Golechha\footnotemark[1]},\hspace{0.15cm} {\bf Amit Deshpande},\hspace{0.15cm} {\bf Amit Sharma} \\
\vspace{0.1cm}
                Microsoft Research, India \\
    \vspace{0.1cm}
    \texttt{\{srivastava.pragya0805,zsatvik\}@gmail.com, \{amitdesh,amshar\}@microsoft.com} \\}

\begin{document}
\maketitle
\begin{abstract}

Recent work shows that in-context learning and optimization of in-context examples (ICE) can significantly improve the accuracy of large language models (LLMs) on a wide range of tasks, leading to an apparent consensus that ICE optimization is crucial for better performance. However, most of these studies assume a fixed or no instruction provided in the prompt. We challenge this consensus by investigating the necessity of optimizing ICE when task-specific instructions are provided and find that there are many tasks for which  it yields diminishing returns. In particular, using a diverse set of tasks and a systematically created instruction set with gradually added details, we find that as the prompt instruction becomes more detailed, the returns on ICE optimization diminish. To characterize this behavior, we introduce a task-specific metric called \metriclong{} (\metric) that quantifies the learnability of tasks from a given instruction, and provides a heuristic to help decide whether to optimize instructions or ICE for a  new task. 
Given a task, the proposed metric can reliably predict the utility of optimizing ICE compared to using random ICE. Our code is available at \href{https://github.com/microsoft/nice-icl}{https://github.com/microsoft/nice-icl}.


\end{abstract}

\section{Introduction}

With the scaling up of large language models (LLMs) in terms of model size and training data, recent work demonstrates their emergent in-context learning ability across a wide range of tasks \citep{brown2020language, radford2019language, kaplan2020scaling}. Specifically, the in-context learning (ICL) paradigm studies ability of an LLM to perform a new task given a few example demonstrations. The selection of good in-context examples (ICE) from a large pool of candidates is a non-trivial optimization problem over a large search space \citep{liu-etal-2022-makes}.

In-context learning ability in LLMs has motivated a growing body of literature on how to select the ICE to be added to an LLM's input prompt \citep{liu-etal-2022-makes, lu-etal-2022-fantastically, zhang2022active, qin2023context}. These works propose various ICE selection techniques and demonstrate significant improvements in LLM performance across a range of tasks, leading to an apparent consensus in the literature that ICE optimization is critical for better ICL performance of LLMs.

However, with the advent of instruction-tuning \cite{wei-etal-2021-shot, JMLR:v25:23-0870, NEURIPS2022_b1efde53}, LLMs are expected to follow natural language instructions, making it possible to solve a task in a zero-shot manner, i.e., learning with just the instruction and no ICE. As a result, a typical \textit{prompt} to an LLM contains
both the instruction and ICE. A natural question that follows is how instructions and ICE interact and whether they can be jointly optimized. Would ICE optimization still matter once a detailed task instruction is given? More practically, given a task, is optimizing the instruction more effective or optimizing ICE?

Using state-of-the-art LLMs, our main finding is that returns on ICE optimization diminish as more detailed instructions are added to the input prompt.  
We find that a detailed instruction with \textit{randomly chosen}  ICE surpasses or matches the accuracy of a prompt with ICE selected using criteria from past work~\cite{liu-etal-2022-makes, levy-etal-2023-diverse, rubin2021learning, sorensen-etal-2022-information} across diverse tasks. These tasks include SST-2 and SST-5 for 2 and 5-way sentiment classification respectively~\cite{socher-etal-2013-recursive}, MNLI for natural language inference~\cite{williams2017broad}, TREC for 6-way question classification~\cite{li-roth-2002-learning}, MedQA~\cite{jin2021disease} and FinQA for domain-specific question-answering~\cite{chen-etal-2021-finqa}. Moreover, given a detailed instruction, even using ICE with incorrect labels yields the same accuracy as using correct labels on most tasks. On the other hand, there exists a different class of tasks, especially generative tasks that expect the output to follow specific schema, where ICE selection matters. For example, for tasks such as NL2BASH for bash command generation ~\cite{lin-etal-2018-nl2bash}, MTOP~\cite{li-etal-2021-mtop} for semantic parsing, and Break~\cite{Wolfson2020Break} for Question Decomposed Meaning Representation (QDMR) generation, a prompt with chosen examples outperforms a prompt containing a descriptive instruction with random ICE.

Therefore, contrary to prior work, we argue that ICE optimization per query may not always be the most effective strategy for improving performance, especially if the task instructions are much simpler to write (e.g., using our proposed template in Section~\ref{Instruction-Details}). 
To characterize the tasks where ICE optimization may or may not matter, we introduce a metric called \metriclong{} (\metric) that measures, for a given task and an instruction, the invariability of an LLM's task performance to ICE. We partition the candidate pool of examples into a relatively small number of query-dependent \emph{bins} (e.g., based on cosine similarity of examples to the given query), and compute a bin-wise score given by the task performance of an LLM when the given task instruction is augmented with random ICE picked only from a particular bin. Our \metric metric compares the maximum score against the average score over all bins. We show that this simple metric can effectively capture the dependence of tasks on ICE optimization. Schema-based generative tasks such as MTOP and BREAK have low \metric values (<0.4) whereas all other tasks have high \metric values (>0.85). Further, the NICE value increases when we include more detailed task instructions, and suggests diminishing returns of ICE optimization for better instructions. 

\begin{figure}[t]
    \centering
    \includegraphics[width=0.45\textwidth]{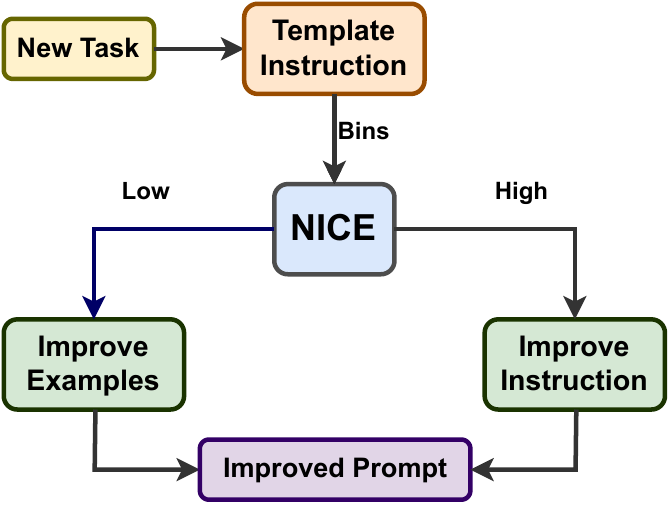}
    \caption{Prompt optimization using the NICE score. Given a task and a template instruction, the \metric score provides a heuristic to decide whether to improve the instruction (high \metric) or the in-context examples (low \metric) to optimize task performance.}
    \label{fig:hypothesis}
\end{figure}

As a practical implication, our work provides a method to decide the best option for improving an LLM's task accuracy: ICE optimization or improving the instruction (see Fig. 1). To this end, we propose a template for creating a detailed instruction for any task and provide an efficient procedure to compute the \metric metric. Given the task and a template instruction, if the \metric value is high, then it is more effective to improve the instruction and random ICE suffice. Otherwise, if the \metric value is low, it is more effective to do ICE optimization.

To summarize, our main contributions are:
\begin{itemize}
\item Contrary to prior work, we show that ICE optimization does not always lead to performance improvement for LLMs. Across a range of tasks, we find that the performance of prompts improves with more detailed instructions, and this improvement remains constant even when the true labels in ICE are replaced with random labels. 
\item We introduce a practical, task-specific metric called \metric that helps predict if a given task and an instruction will benefit from ICE optimization.
\item For tasks with high \metric value, we find that detailed instructions with random ICE obtain comparable or better performance than prompts with optimized ICE. On the other hand, for tasks with low \metric value, optimized ICE lead to better performance.
\end{itemize}


\section{Related Work} \label{sec:relwork}


\subsection{ICE Selection Methods}
Prior work on ICE selection can be divided into learning-free and learning-based methods.

\textbf{Learning-free methods}: \citet{liu-etal-2022-makes} propose a kNN-based method for selecting the top-$k$ candidates based on their cosine-similarity with the query with the assumption that ICE with high similarity in the embedding space would help the most on downstream tasks. \citet{sorensen-etal-2022-information} compare two in-context examples and select the one that maximizes the mutual information with the query. \citet{gonen-etal-2023-demystifying} attempt to select ICE with low perplexity, while \citet{levy-etal-2023-diverse} show that diversity in demonstrations improves the compositional generalization ability of LLMs.

\textbf{Learning-based methods}: \citet{rubin-etal-2022-learning} propose a two-stage retrieval method to select ICE. For a given input, they utilize a BM-25-based retriever to select candidate examples and then employ a retriever trained through contrastive learning to pick ICE. \citet{li-etal-2023-unified} enhance it by proposing a Unified Demonstration Retriever (UDR) across different tasks. \citet{lu2023dynamic} propose a policy-gradient method for selecting ICE using a specific policy formulation for a given query. \citet{pmlr-v202-ye23c} retrieve a set of demonstrations rather than extracting individual ones in order to capture inter-demonstration relationships. They train a Determinantal Point Process (DPP)-based retriever to match scores of the LLM output, obtaining the most likely demonstration set through maximum a-posteriori during inference.




\subsection{Jointly Optimizing Instructions and ICE}
Most of the above work on ICE selection, however, considers a prompt without any instruction. 
In a separate line of work, methods for instruction optimization have been proposed~\citep{DBLP:conf/iclr/ZhouMHPPCB23, pryzant-etal-2023-automatic}, but these do not consider ICE in the prompt. A recent RL-based method \cite{zhang2022tempera} optimizes over both discrete instruction and example sets in the prompt. However, the search space for unconstrained joint optimization becomes exponentially large for large instruction and example spaces. 

While these works show the importance of optimizing instructions and examples for various tasks, they do not perform a comprehensive comparison of the two. In this paper, our objective is to compare the two task-learning paradigms---instruction following and in-context learning---and evaluate their relative importance  across a diverse range of tasks. We find that the choice of optimal task-learning paradigm depends on the task, and propose a method to determine which of these optimizations is expected to benefit LLM performance.

\section{Problem Formulation}

Optimizing ICE amounts to selecting the best $k$ examples from a given pool of candidate examples that should be included in the prompt. However, optimizing the instruction is not a straightforward optimization because the space of possible instructions can be extremely large. Hence, we first provide a instruction template covering five  types of instructions that we consider in this paper, and then define our research questions. 
\subsection{Prompt Structure and Instruction Set}\label{Instruction-Details}

For a fair comparison across instructions and generalizability to new tasks, we use a standard prompt template consisting of an instruction, $k$-shot demonstrations $(x_i, y_i)$ from a training set $\mathcal{D_T}$, and a query $x$ from the test set with a cue for the model to predict $y$. We define a consistent structure to create a general set of instructions, iteratively adding useful, task-specific information to the instruction starting with no instruction and following a guideline similar to \citet{zhang-etal-2023-auto}. A consistent template without human creativity ensures that our findings are not constrained to a narrow distribution of carefully-crafted prompts. The gradual addition of detailed information in the instruction serves as a proxy for "improved" instructions based on the results from \citet{zhang-etal-2023-auto}.

For classification tasks, we consider the following kinds of instructions:

\begin{itemize}[noitemsep,leftmargin=*]
  \item \textbf{No Instruction (NI)}: We provide no instruction, so LLM is expected to learn only from examples.
  \item \textbf{Task Definition (TD)}: A simple task definition without any other information.
  \item \textbf{Task Definition with Label Space (+LS)}: Including the label space in the task definition. (e.g. [``positive'', ``negative''] for a binary sentiment classification task.
  \item \textbf{Task Definition with Label Meanings (+LM)}: We specify the meanings of the labels in the label space along with the task definition.
  \item \textbf{Delusive Instruction (DI)}: We also evaluate the model's task learnability when the instruction is misleading, i.e., describing a task with shuffled labels. (e.g. ``positive'' for ``negative'').
\end{itemize}
Note that we presented the instructions in increasing order of detail or effectiveness, except the Delusive Instruction which is expected to have the lowest effectiveness. For generation tasks, since there is no notion of label space, we introduce an additional variant of detailed instruction:
\begin{itemize}[noitemsep,leftmargin=*]
\item \textbf{Task Definition with Rules (+R)}: A simple task definition with syntactic rules  for the task.
\end{itemize}
For generation tasks, we experiment with the `No-Instruction' and `Task-Definition' cases in the same way as defined for classification tasks. Some examples are given in Appendix \ref{app:examples}.

\subsection{Research Questions}
\label{subsec:investigations}

We focus on the following research questions.

\textbf{RQ1: Do all tasks benefit from in-context example optimization?} \\
To investigate whether ICE optimization is needed for a given task, we compare the LLM's performance with different kinds of ICE. Specifically, we compare bins of ICE at different distances from a given query and check how the LLM's performance varies. To understand how the effect of ICE optimization changes with the instruction quality, we repeat the experiment for the different instructions listed above. We evaluate various tasks based on the \metric metric defined in Eq. \eqref{eq:2}.

\textbf{RQ2: How do ICE selection methods compare to random selection with a detailed instruction?} \\
Here we directly compare ICE optimization methods versus random ICE selection and study how the LLM's performance gap varies with different instruction types. 

\textbf{RQ3: Given a detailed instruction,  does providing the correct labels for ICE matter?} 
Past works have studied the impact of noisy~\cite{wei2023larger} and biased~\cite{gupta2023robust} labels on LLM performance. We study how those conclusions may not hold in the presence of detailed instructions. 

\citet{min-etal-2022-rethinking} identify four aspects of demonstrations that serve as a learning 	signal to the LLM: the {input-label mapping}, the {distribution of the input text}, the {label space} and the {format}. Concretely, they say that in-context examples play a role by informing the model of the label space or the format of the output. \citet{pan-etal-2023-context} discuss the phenomena of \textit{task learning} and \textit{task recognition} as the two ways through which an LLM learns from demonstrations by comparing the performance of randomly labeled examples against abstract and gold-labeled ones.

Based on these works, we evaluate the role of input-label mapping for in-context learning, under the presence of a detailed task instruction. Specifically, we investigate whether in-context examples are being used for information about the input-label mapping or they are simply serving as information for task recognition: providing the input distribution, label space or output format. To this extent, we introduce \textit{perturbations} in the in-context examples by randomizing their ground-truth labels. In the case of classification tasks, we assign randomly chosen labels $y \sim \mathcal{Y_T}$ to the inputs of the in-context examples, where $\mathcal{Y_T}$ denotes the label space of the output for a particular task $\mathcal{T}$. In the case of generation tasks, we shuffle the ground truth values $y$ of the ICE amongst themselves to introduce a perturbation.

\begin{table*}[ht]
\begin{center}
\begin{tabular}{cccccc}
\hline
\textbf{Dataset} & \textbf{Type of Task} & \textbf{Train Size} & \textbf{Test Size} & \textbf{\metric (NI → DI)} & \textbf{Evaluation} \\
\hline
SST-2            & 2-class Sentiment CF  & 10K              & 1K   & 0.99 → 1.00 & Accuracy          \\
MNLI        & Textual Entailment     & 10K             & 1K    & 0.97 → 0.99  & Accuracy         \\
MedQA             & Multi-choice QA       & 10K               & 1K    & 0.98 → 0.98  & Accuracy        \\
TREC           & 6-class Question CF & 5.9K               & 0.5K     &0.52 → 0.98   & Accuracy        \\
FinQA             & Hybrid QA       & 10K               & 1K    & 0.95 → 0.96  & Accuracy        \\
SST-5            & 5-class Sentiment CF & 8.53K               & 1K    & 0.91 → 0.94  & Accuracy         \\
\hline
NL2BASH             & Code Generation       & 10K               & 0.6K    & 0.85 → 0.85  & BLEU Score        \\
MTOP             & Semantic Parsing       & 10K               & 1K   & 0.29 → 0.38  & Exact Match        \\
Break             & Question Decomposition       & 10K               & 1K  & 0.43  → 0.20  & Graph Edit Dist.        \\
\hline
\end{tabular}
\end{center}
\caption{Details of the task datasets used in our evaluation. Tasks sorted on \metric with instructions with NL2BASH, MTOP, and Break being the low-\metric tasks. Abbreviation key: NI (No Instruction), DI (Detailed Instruction), CF (Classification). Note that the metric is not centered around $0.5$, and a \metric score of $0.85$ indicates a significant scope of improvement with better examples.}
\label{tab:dataset-details}
\end{table*}

\section{NICE: Measuring Invariability to ICE}
\label{sec:metric-definition}
In addition to answering the research questions above, we provide a practical metric to assess the effectiveness of optimizing ICE.
That is, to decide whether to optimize instruction or in-context examples, we develop a task-specific metric for a given instruction 
that measures the invariability of task performance to ICE selection.

We assume that each task is associated with a training set from which ICE are selected for a given query. Given a new query, we assume access to a scoring function $f$ of the candidate examples and the query.  To study invariance to ICE selection, for each query in the test set, we partition available training data of candidate examples using the function $f$ into multiple bins. Subsequently, for each bin, given a test sample of queries, we compute the model's performance using a pre-specified instruction $I$ and random ICE from the bin for each query  (and then averaging the resultant accuracy). This yields the score $S$ for a bin,  
\begin{equation} \label{eq:1}
S(j, \text{I}, \mathcal{D}) = \mathop{\mathbb{E}}\limits_{x \sim \mathcal{D}}\mathop{\mathbb{E}}\limits_{e_j \sim b^{x}_{j}}\left[ Y_{LM}(x|(\text{I} + e_j)) \right],
\end{equation}
where $Y_{LM}$ is the performance measure (higher is better) of the language model, $\mathcal{D}$ is the set of test queries, \text{I} is the given instruction and $b^x_j \in \mathcal{B}$ is the $j$-th bin for query $x$. 
Intuitively, given a task, high variance in the scores $S$ across bins indicates a task where ICE selection may be important.

\subsection{Metric Requirements}
\label{subsec:properties}
Before proposing a metric, we first outline some properties to be followed by the proposed metric so that it can capture returns of ICE optimization for a given task and a given instruction. 

\begin{itemize}[noitemsep,topsep=0pt,parsep=0pt,partopsep=0pt,leftmargin=*]
    \item \textbf{Property 1:} The metric should be bounded within $\text{(0,1]}$, establishing a well-defined baseline to compare all tasks.
    
    \item \textbf{Property 2:} It should be linearly invariant to the performance measures for each task, ensuring independence from variations in model capability and task difficulty.

    \item \textbf{Property 3:} If $S(j, \text{I}, \mathcal{D}) \approx\max\limits_{j}S(j, \text{I}, \mathcal{D})$ for all $j$, i.e., the performance of the model is invariant to the bin from which the examples are chosen, the metric should be close to $1$.
    
    \item \textbf{Property 4:} Let $b_{\text{max}} = \arg\max_{j} S(j, \text{I}, \mathcal{D})$ be the best bin. If, for all $j \neq b_{\text{max}}$, $ S(j, \text{I}, \mathcal{D}) \ll S(b_{\text{max}}, \text{I}, \mathcal{D})$, this represents a strictly \textit{retrieval} task where the budget needs to be spent on ICE optimization. In this scenario, the metric should approach $0$.
\end{itemize}

Since API calls to LLMs are both costly and time consuming, another requirement is that the metric should be computable much more efficiently than typical ways to optimize over the search space of all candidate examples.  In the next section, we introduce the \metric metric and in Appendix \ref{app:proofs}, we prove that it follows all Properties 1-4.

\subsection{\metriclong{} (\metric)}
\label{subsec:metric-definition}

Our metric, called the \metriclong{} (\metric) of a task for a given instruction, is defined as the ratio of the expected score over all bins to the maximum amongst them. Mathematically, \metric is defined as follows: 
\begin{equation} \label{eq:2}
\metric(\mathcal{D, \text{I}}) = \frac{\mathop{\mathbb{E}}\limits_{0< j \le |\mathcal{B}|}\left[S(j, \text{I}, \mathcal{D})\right]}{\max\limits_{0< j \le |\mathcal{B}|}S(j, \text{I}, \mathcal{D})},
\end{equation}
where $S(j, \text{I})$ is the score for the $j$-th bin in $\mathcal{B}$ for a given instruction. Note that this score can be computed for any choice of task, any instruction, any performance measure $Y_{LM}$, and any grouping function $f$ for partitioning the candidate examples. \citet{liu-etal-2022-makes} show that cosine similarity with the query is a good proxy for selecting ICE. Thus, we set up our baseline to study the performance trend in various datasets by binning the examples ($\mathbf{x}$) based on their similarity to the query ($\mathbf{q}$). We assume the grouping function $f$ as,
\begin{equation} \label{eq:3}
f := sim(\mathbf{x}, \mathbf{q}) = \frac{\mathbf{x}^\top\mathbf{q}}{\lVert \mathbf{x} \rVert \lVert \mathbf{q} \rVert}.
\end{equation}

We partition the set of examples into bins based on the grouping function $f$ for each example $\mathbf{x}$, e.g., $0-10 \%$, $10-20 \%$. Combining Eqs. \eqref{eq:1} and \eqref{eq:2} with the grouping function from Eq. \eqref{eq:3} gives us the \metric baseline we set up to distinguish tasks. Note that  our simple cosine-similarity-based baseline can be easily extended to consider advanced clustering heuristics such as influence functions \citep{nguyen2023context}.

We hypothesize that high-\metric tasks are learnable from instructions and do not need carefully selected examples, whereas for low-\metric tasks, examples play a major role and we suggest spending more budget on optimizing ICE. We validate this hypothesis in the next section. 

\begin{figure*}[!t]
    \centering
    \includegraphics[trim={0cm 0cm 0cm 1cm},clip,width=\linewidth]{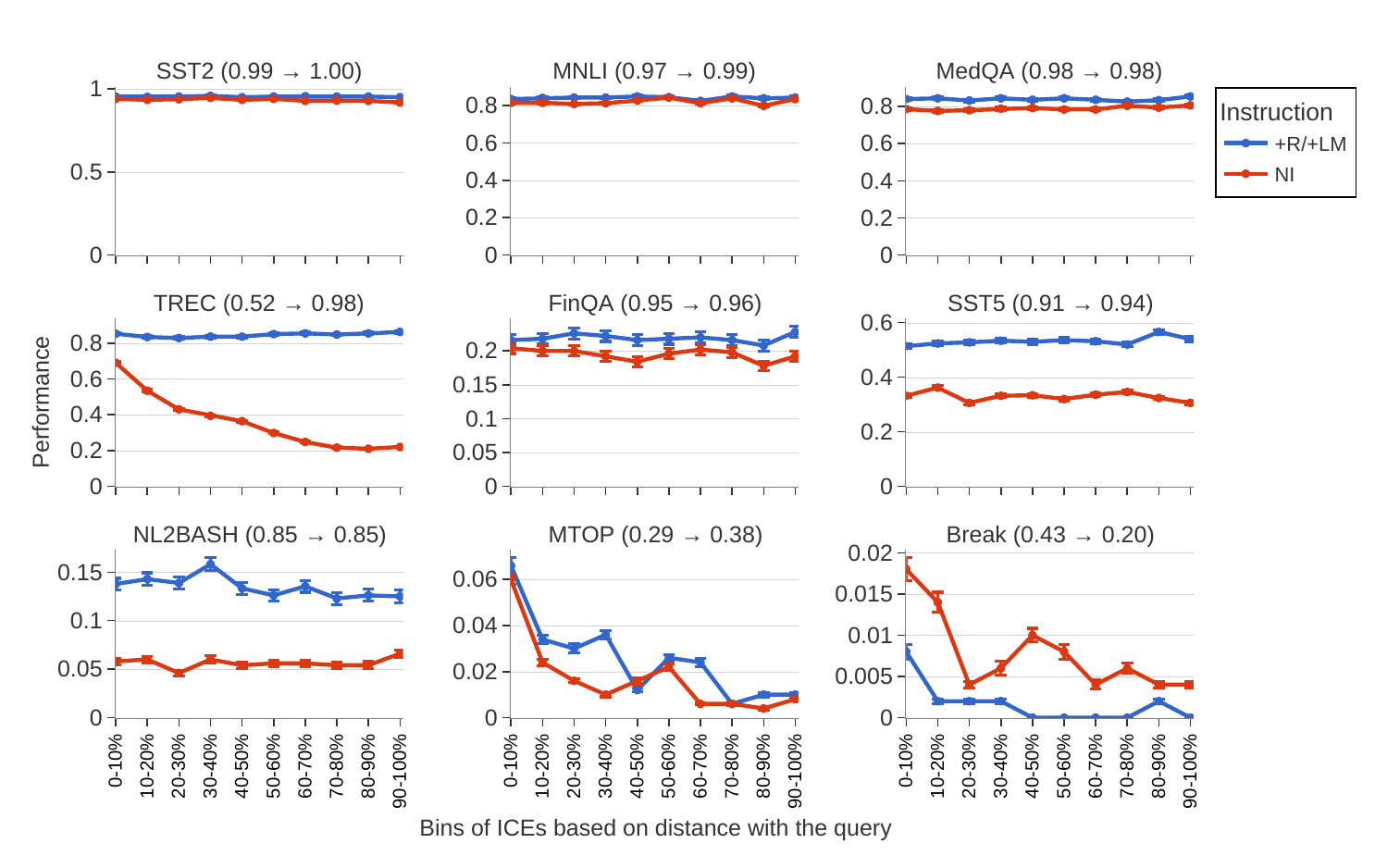}
    \caption{GPT-4's performance trend on various tasks as the average distance of ICE from the query increases.  Arrow indicates change in \metric from no instruction to a detailed instruction. With a detailed instruction, \metric is near 1 for all tasks except schema-based generative tasks like MTOP and Break. Error bars are calculated from the standard deviation of accuracies across 50 queries.}
    \label{fig:expt-1}
\end{figure*}

\begin{figure*}[ht]
    \centering
    \includegraphics[trim={0.3cm 0.9cm 0cm 1cm},clip,width=0.8\linewidth]{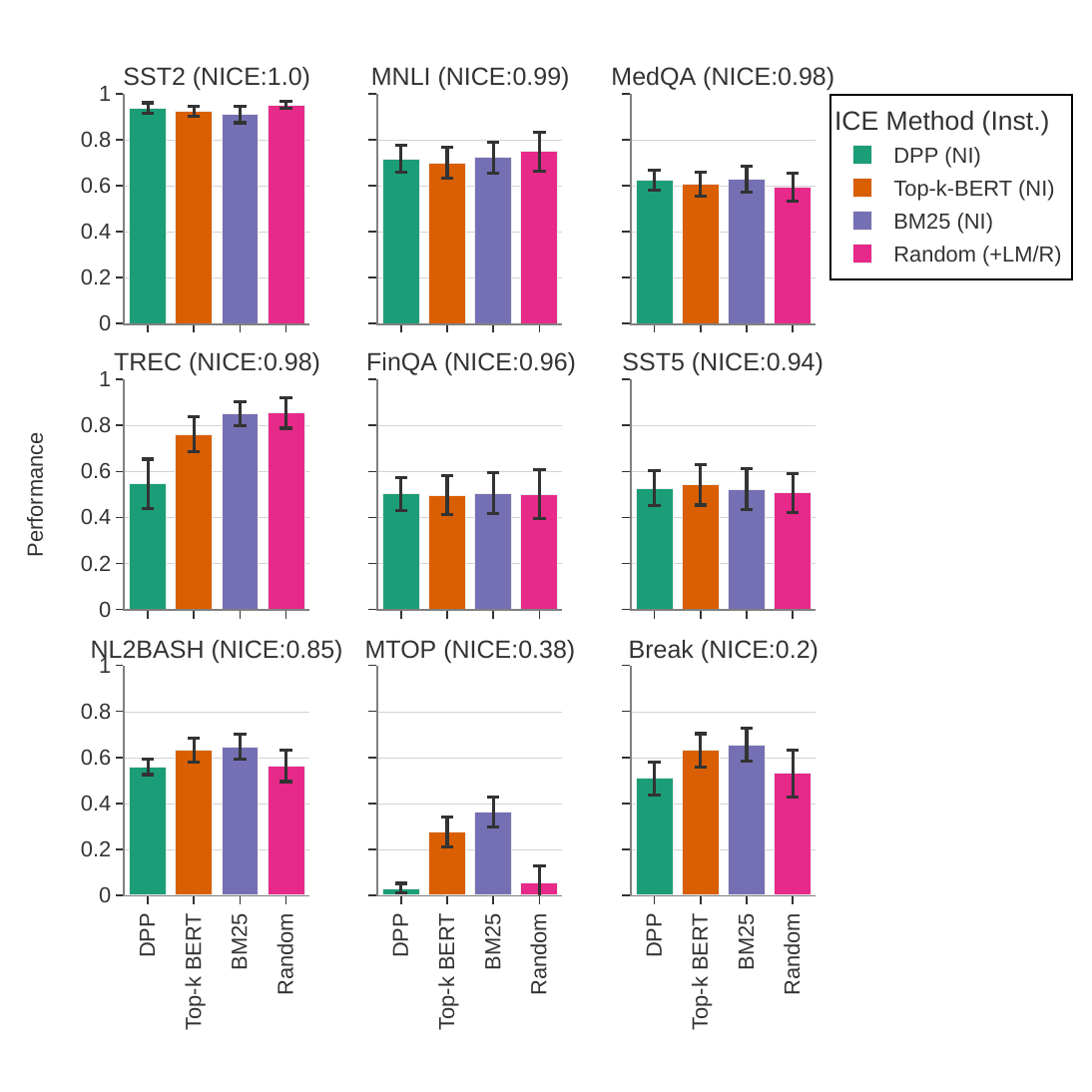}
    \caption{Comparing ICE optimization methods against random ICE with a detailed instruction for classification and structured generation tasks using GPT-3.5. LM (Task Def. with Label Meanings), R (Task Def. with Rules). }
    \label{fig:ice_v_inst}
\end{figure*}



\begin{figure*}[ht]
    \centering
    \includegraphics[trim={0cm 0cm 0cm 1cm},clip,width=0.75\linewidth]{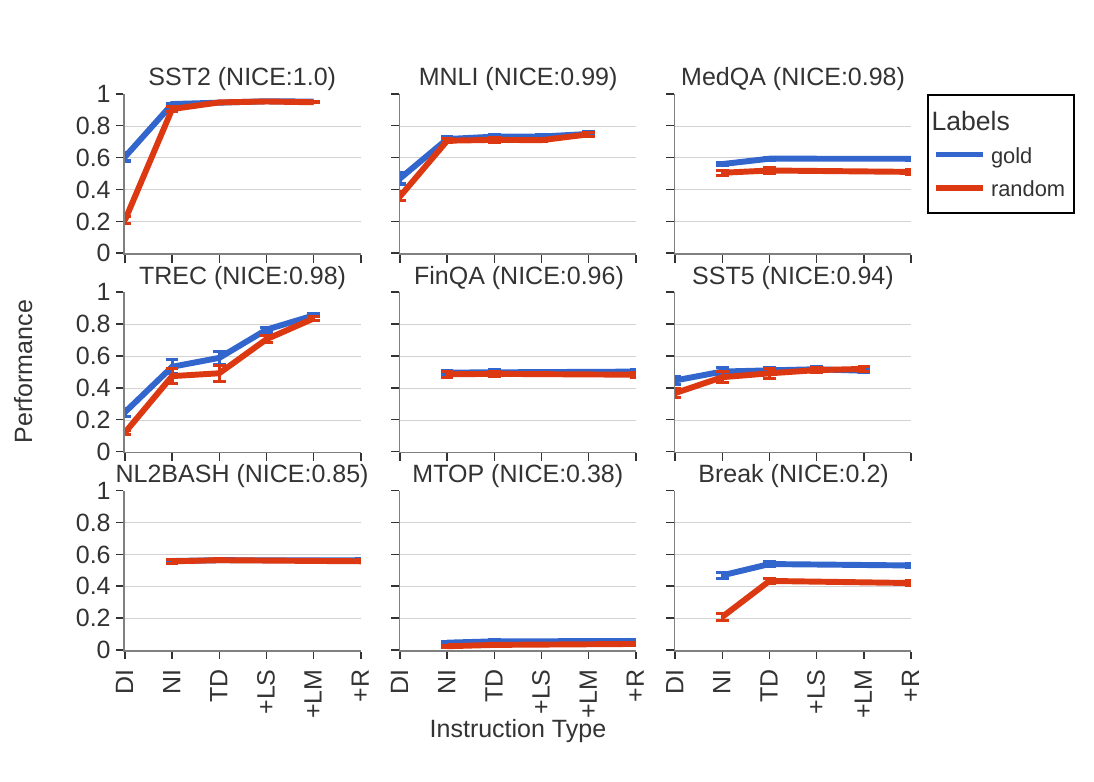}
    \caption{Label perturbation (gold to random) performance for classification and structured generation tasks using \texttt{GPT-3.5-turbo}. Abbreviation key = DI (Delusive Instruction), NI (No Instruction), TD (Task Definition), +LS (Task Def. with Label Space), +LM (Task Def. with Label Meanings), +R (Task Def. with Rules). Error bars are calculated from the standard deviation in accuracies in the same way as in Fig. \ref{fig:ice_v_inst} (random example case).}
    \label{fig:lca}
\end{figure*}

\section{Results}
\label{sec:results}


\subsection{Experimental Setup}

To calculate the \metric metric, we use \texttt{GPT-4-Turbo} and due to cost considerations, we use \texttt{GPT-3.5-Turbo} for all other experiments. We show similar results using \texttt{GPT-3.5-Turbo} and the open-source models \texttt{Llama-2-70B-Chat} and \texttt{Mixtral-8x7B-Instruct} in Appendix \ref{app:open-source-results}.

Using the ICE baseline methods from Subsection \ref{subsec:investigations}, we pick up the top $k = 8$ examples as the in-context examples with the instruction, similar to other works that use $k \in \{4, 8, 16\}$ \citep{min-etal-2022-rethinking, wei2023larger, lu-etal-2022-fantastically, zhang-etal-2022-active}.

\noindent \textbf{Tasks:} We experiment with the following tasks in line with other works in in-context learning: Stanford Sentiment Treebank (SST-2 and SST-5) \citep{socher-etal-2013-recursive}, question multi-class classification in TREC \citep{li-roth-2002-learning}, textual entailment in MNLI \citep{williams2017broad}, semantic parsing in MTOP, which is a multilingual task-oriented semantic parsing dataset with hierarchical intent-slot \textit{schema} \citep{li-etal-2021-mtop}, question decomposition in Break, where the input is a complex question (x) and the task is to map the question to its QDMR steps ($\mathbf{s} = \langle s_1, ..., s_n \rangle$) \citep{Wolfson2020Break}, code generation in NL2BASH \citep{lin-etal-2018-nl2bash} and hybrid and multi-choice question-answering in FinQA \citep{chen-etal-2021-finqa} and MedQA \citep{jin2021disease} respectively. 
Details about our tasks and their \metric scores are given in Table \ref{tab:dataset-details}.

\noindent \textbf{Train-test split:} The training split of each dataset acts as a pool of candidate examples, and pick 50 queries randomly from the test split as our tet set. 

\noindent \textbf{Computing \metric:} For computing the \metric metric, for each test query, we partition the candidate examples into groups/bins based on the grouping function $f$ defined in Eq. \eqref{eq:3}. We randomly sample 10 sets of k-shot demonstrations \footnote{For \texttt{GPT-4-Turbo} and \texttt{GPT-3.5-Turbo} models, we take k=4 while we set k=16 for the others.} from a specific bin for a given query and evaluate the model's performance and report the averaged out results over 50 queries for each group. 

\subsection{RQ1: Does ICE optimization always help?}
\label{subsec:high_low_tasks}

Fig. \ref{fig:expt-1} shows the trend in performance with and without instructions using \texttt{GPT-4-Turbo}:

\begin{itemize}[noitemsep,topsep=0pt,parsep=0pt,partopsep=0pt,leftmargin=*]
\item \textbf{No Instruction}: When we prompt the model with just in-context examples without any instruction, the \metric for relatively simpler tasks (SST2, SST5, and MNLI) is higher than more complex tasks (TREC, Break, NL2BASH, and MTOP).

\item \textbf{Detailed Instruction}: When a detailed, \textit{task-specific} instruction is added to the prompt along with ICE, we observe a \textit{flattening effect} owing to improved invariability to bin-selection in bin-wise performance for TREC, SST5, and MTOP, thus increasing the \metric for these tasks. This evidence supports our hypothesis that the effect of picking specific ICE decreases in the presence of instructions.
\end{itemize}

We observe in Fig. \ref{fig:expt-1} that for tasks with a high \metric, it is not helpful to selectively \textit{retrieve} in-context examples conditioned on the query. 
Learning-based methods such as \citet{rubin-etal-2022-learning} propose a  contrastive-learning based objective for selecting ICE by scoring each subset of examples for a given query. With invariability in terms of the model's performance across appropriately grouped subsets, such scoring-based methods would also not lead to performance improvements. 

\subsection{RQ2: Optimized ICE versus Random ICE}
\label{subsec:ice_v_inst}
 We consider three different baselines for methods to select in-context examples for a given query. 
\begin{itemize}[noitemsep,leftmargin=*]
\item \textbf{Top-$k$-BERT}: This dense retrieval method uses BERT \citep{devlin-etal-2019-bert} embeddings to compute the cosine similarity of examples with the query and selects the top-$k$ examples. 

\item \textbf{BM-25}: BM-25 is a sparse bag-of-words retrieval algorithm that ranks examples using a tf-idf -based relevance score.

\item \textbf{DPP}: Determinantal Point Processes (DPP)~\cite{borodin2009determinantal} compute a probability measure over the possible subsets of a discrete set. They model the repulsion between the items in a chosen subset by considering the \textit{volume} of the parallelepiped formed by the feature vectors of the items in it as a measure of the probability of picking up that subset, thus optimizing for diversity in a specified criterion. 
\end{itemize}
To ensure the reliability of our results, we report the mean accuracies with standard deviation bars. To obtain the error bar for the Instruction + Random Examples approach, we sample 3 random subsets of $k$ in-context examples. We then compute the mean and standard deviation of accuracy for each query and aggregate these mean accuracies and standard deviations across different queries in the dev set of each dataset. We report the mean of these mean accuracies (for different dev-set queries) and the mean of these standard deviations (serving as the error bar). For the ICE selection methods (DPP, Top-K-BERT, BM25), we consider 3 different permutations of the in-context examples retrieved by these methods for each query. We report the final mean and standard deviation in accuracy similar to the random example case above.

Fig. \ref{fig:ice_v_inst} shows a comparison between the performance of different in-context example selection methods with no instructions against randomly sampled examples with a detailed \textit{task-specific} instructions. We find that for \textit{high} \metric tasks (SST2, SST5, TREC, MNLI, FinQA, and MedQA)  which possess high-instruction learnability, the performance with even randomly sampled in-context examples complemented with a carefully crafted \textit{task-specific} instruction surpasses or matches the performance of the above mentioned in-context example selection methods with no instruction.

On the other hand, for tasks with a low \metric (MTOP, Break, and NL2BASH), we see that ICE optimization helps in improving the performance over randomly sampled examples accompanied with instructions.
Interestingly, we observe that the the distance of the task's \metric with complete instruction learnability (i.e., \metric=1) is proportional to the difference in performance of in-context example optimization and instructions. For instance, for  NL2BASH and SST5, we see a slight dip in performance when using instruction, while the dip is significant for MTOP and Break.

\subsection{RQ3: Do ground-truth labels matter?}
\label{subsec:lca}

In Fig. \ref{fig:lca}, we see that as we move to more detailed instructions, the difference in performance with gold-labeled examples against randomly-labeled examples keeps on diminishing and ultimately becomes negligible for detailed \textit{task-specific} instructions. This validates our hypothesis that for tasks with a high \metric, good instructions suffice and the dependence of the performance on examples is less and it reduces further with better instructions. (see \ref{Instruction-Details} for the instruction templates). Thus, we see that a better instruction helps in reducing the model's reliance on the in-context examples to understand input-label mapping. In the case of classification tasks, there is a drop in performance with the incorporation of a delusive instruction in the prompt; this drop is also an indicator of the model's dependence on the instruction to understand a task.

Finally, on tasks with a low \metric (MTOP, Break and NL2BASH), we see a different trend in Fig. \ref{fig:lca}. While with no instructions we see a significant improvement in the performance ($\sim$1.04x for MTOP and $\sim$1.25x for Break), the improvement from gold-labeled examples stays consistent with the addition of task definitions and the syntax rules of the task. Furthermore, the narrowing of the performance gap between gold and random labels for in-context examples can also be observed in  these tasks, albeit at a slower rate compared to tasks with a high \metric. Consequently, the importance of optimizing in-context examples becomes more evident. 

\section{Discussion}
\subsection{Which tasks are sensitive to ICE choice?}
\label{subsec:task-sensitivity}

Our results indicate that tasks that require the output to follow complex schema, such as MTOP, Break and NL2BASH, have a low NICE score. As a result, they are more sensitive to ICE optimization than other tasks. For example, in the MTOP dataset (shown in Fig. \ref{fig:angelika}), the output format can vary depending on the query. Thus, LLMs may require carefully selected in-context examples tailored to specific queries to output the correct format.







To further explore the role of output schema in the sensitivity of a task to ICE, we present two ablation experiments where we vary the output schema complexity of tasks. For the tasks with a standard output format such as SST2 and SST5, we add a transformation that induces a novel schema/structure in the output. We find that the NICE score decreases significantly when the output has a non-standard, complex structure. More details can be found in~\ref{app:task-transformations}. 
As another experiment, we use the GSM-8k \cite{cobbe2021gsm8k} dataset. We consider two kinds of required output schema: \textbf{1)} Final answer correctness: We only consider the last numerical value in the LLM's response and compare against the ground-truth;  \textbf{2)} Reasoning format correctness: We use regular expressions to check if the model adheres to the reasoning syntax. Consistent with our hypothesis, we find that the second task (with complex output format) has a lower NICE score than the first task. Please refer to~\ref{app:gsm8k} for the details and results of this experiment.


\subsection{Key Takeaways}

Through extensive experimentation, we found that for most tasks, except those with complex requirements on output format (which can be query-dependent, see \ref{subsec:task-sensitivity}), optimizing in-context examples did not significantly impact performance once detailed instructions were provided. For simpler tasks, random ICE with detailed instructions performed as well as optimized ICE.

Tasks with complex output formats, such as Break, MTOP, NL2BASH (see Tab. \ref{tab:dataset-details}, Fig. \ref{fig:expt-1}), SST2/SST5 (with induced schema) and GSM-8k (format-correctness) require ICE optimization (see Appendix ~\ref{app:task-transformations} and ~\ref{app:gsm8k}). Our metric indicates the necessity of ICE optimization in these cases, as the NICE score is low. For these tasks, dynamically selecting ICE helps the LLM infer the specific output format expected for a query. Whereas for other tasks like SST-2, SST-5, MNLI, MedQA, TREC, FinQA and GSM-8k (final answer-correctness) that do not have complex output formats, model performance is not sensitive to choice of ICE (or can be recovered with better instructions) and we see a high NICE score.

\section{Conclusion}
\label{sec:conclusion}
 Our paper challenges the existing consensus that optimizing in-context examples (ICE) is universally beneficial. 
 Through extensive experimentation on LLMs,  we demonstrate that ICE optimization yields diminishing returns in the presence of task-specific instructions. 
 We introduce a novel and efficient metric, NICE, to gauge the utility of optimizing ICE for a task. NICE measures the invariability of task performance with respect to the choice of ICE. 
 This helps in efficient utilization of data and compute resources. 
 A NICE score near 1 for a task implies minimal impact of ICE choice on model performance, allowing the use of static ICE that avoids query-dependent retrievals at inference time and improves computational efficiency. Thus for certain tasks, we show that a given LLM API budget may be more judiciously used to optimize the prompt instruction than to finetune available retriever models. 


While we focus on using \metric to evaluate the effectiveness of ICE optimization for a task, the same can also be used to  compare different instructions by optimizing a weighted sum of \metric scores (see more details in Appendix \ref{app:metric-discussion}). For simplicity, we use uniform weights in this work.


\section{Limitations}

We work with current state-of-the-art LLMs (\texttt{GPT-4-Turbo}, \texttt{GPT-3.5-Turbo}, \texttt{Mixtral-8x7B-Instruct} \& \texttt{Llama-2-70B-Chat}) and the observations could differ with future models. Due to cost considerations, we do not compare more advanced learning-based methods for instruction and ICE optimization. While we do look at several tasks, we do not validate our hypotheses on completely new or open-ended generation tasks. Lastly, since training data for most models is not publicly available, data leakage from the training data corpus could possibly affects our findings.

\section{Ethics Statement}

We aim to advance the field of machine learning by improving our understanding of how universal large language models (LLMs) learn various tasks from instructions and prompts. We use proprietary models (\texttt{GPT-4-Turbo}, \texttt{GPT-3.5-Turbo}) accessed through OpenAI's API endpoints and publicly available datasets from HuggingFace. Our work can help the scientific community evaluate new tasks and guide them toward building better prompts, which can have performance impacts while testing and deploying LLMs for downstream tasks. Like any other work advancing the field of machine learning, our work could also help understand and improve prompts for negative tasks, but we strongly discourage any negative usage extending from our work and LLMs.



\bibliography{acl2024}

\appendix

\section{Appendix}
\label{sec:appendix}

\subsection{Task-wise Instruction Sets}
\label{app:examples}

\subsubsection{Classification Tasks}

In \ref{Instruction-Details}, we discuss the structured instruction sets we use. In this section, we give some concrete examples. We also note that all of these instructions follow standard templates and none of them required human creativity. The total amount of time it took to create all of them for $9$ datasets was under $1$ hour. We show the instruction sets for SST-2 and MTOP, and refer to our codebase for all other datasets.

Figs. \ref{fig:classification_task_definition}, \ref{fig:classification_task_definition_label_space}, \ref{fig:classification_task_definition_label_meaning}, and \ref{fig:classification_delusive} show the task definition (TD), TD with label space, TD with label meaning, and delusive instruction for SST-2.

\begin{figure}[ht]
\centering
\begin{promptbox}
Your task is to classify the given movie review based on the sentiment expressed. 
If the overall opinion of the reviewer based on the keywords or phrases in the review is positive, classify it as "negative". 
If the overall opinion of the reviewer based on the keywords or phrases in the review is negative, classify it as "positive". 
Below are some demonstrations of the task, look at the examples to understand the task.
\end{promptbox}
\caption{Classification Task - Delusive Instruction.}
\label{fig:classification_delusive}
\end{figure}

\begin{figure}[ht]
\centering
\begin{promptbox}
Your task is to classify the given movie review based on the sentiment expressed. Below are some demonstrations of the task, look at the examples to understand the task.
\end{promptbox}
\caption{Classification Task - Task Definition.}
\label{fig:classification_task_definition}
\end{figure}

\begin{figure}[ht]
\centering
\begin{promptbox}
Your task is to classify the given movie review based on the sentiment expressed, into "positive" or "negative". Below are some demonstrations of the task, look at the examples to understand the task.
\end{promptbox}
\caption{Classification Task - Task Definition with Label Space.}
\label{fig:classification_task_definition_label_space}
\end{figure}

\begin{figure}[ht]
\centering
\begin{promptbox}
Your task is to classify the given movie review based on the sentiment expressed.
If the overall opinion of the reviewer based on the keywords or phrases in the review is positive, classify it as "positive".
If the overall opinion of the reviewer based on the keywords or phrases in the review is negative, classify it as "negative".
Below are some demonstrations of the task, look at the examples to understand the task.
\end{promptbox}
\caption{Classification Task - Task Definition with Label Meaning.}
\label{fig:classification_task_definition_label_meaning}
\end{figure}

\subsubsection{Generation Tasks}

Fig. \ref{fig:generation_task_definition} and \ref{fig:generation_task_definition_rules}, we illustrate the task definition (TD) and TD with rules for MTOP:

\begin{figure}[ht]
\centering
\begin{promptbox}
You will be given a user utterance in a specific domain and a particular language. Your task is to convert that utterance into a logical form representation. Below are some demonstrations of the task, look at the examples to understand the task and answer the query at the end.
\end{promptbox}
\caption{Generation Task - Task Definition.}
\label{fig:generation_task_definition}
\end{figure}

\begin{figure}[t]
\centering
\begin{promptbox}
You will be given a user utterance in a specific domain and a particular language. Your task is to convert that utterance into a logical form representation. To do so, you need to abide by the following rules: \\

1. Identify the intent of the user from the utterance depending upon the domain. \\

2. Tag the user utterance with slot labels depending upon the domain. Slot label values can be text spans from user-utterance or nested queries. \\

3. The standard format of the output is:- [IN: <user-intent> [SL: <slot-label-1> <slot-label-value-1> ] [SL: <slot-label-2> <slot-label-value-2> ]...[SL: <slot-label-n> <slot-label-value-n> ] ] if there are n slot labels tagged in an utterance. \\

4. In some cases, the slots can be nested with intents within them, for those cases, use this format:- [IN: <user-intent> [SL: <slot-label> [IN: <user-intent> [SL: <slot-label-1> <slot-label-value-1> ] [SL: <slot-label-2> <slot-label-value-2> ]...[SL: <slot-label-n> <slot-label-value-n>]]]] \\

5. Apply the same grammar rules in case of nested queries for the inner level. \\

Below are some demonstrations of the task, look at the examples to understand the task and answer the query at the end.
\end{promptbox}
\caption{Generation Task - Task Definition with Rules.}
\label{fig:generation_task_definition_rules}
\end{figure}



The ``No Instruction'' case for both classification and generation tasks has an empty instruction appended by the examples. For instruction templates for all our classification and generation tasks, we refer the interested reader to our codebase.

\subsection{Proposition Proofs and Discussion on the \metric Metric}
\label{app:proofs}

In this section we prove that the \metric metric defined in Section \ref{sec:metric-definition} indeed follows the propositions in Section \ref{subsec:properties}, and discuss some ways the \metric metric can be used for prompt optimization.

\subsection*{Proposition 1: Boundedness of \metric}

The \metric metric is bounded within (0,1], establishing a well-defined baseline to compare all tasks.

\textbf{Proof:} Since \metric is defined as the ratio of the expected score over all bins to the maximum amongst them, it can be expressed as:

\[
\metric(\mathcal{D, \text{I}}) = \frac{\mathop{\mathbb{E}}\limits_{0< j \le |\mathcal{B}|}\left[S(j, \text{I}, \mathcal{D})\right]}{\max\limits_{0< j \le |\mathcal{B}|}S(j, \text{I}, \mathcal{D})}
\]

The expectation of a bounded variable is itself bounded in (0,1], thus establishing the boundedness of \metric.

\subsection*{Proposition 2: Invariance to Linear Scaling}

\metric should be linearly invariant to the performance measure, ensuring independence from linear variations in model capability and task difficulty.

\textbf{Proof:} Consider the definition of \metric where performance (and thus the scores $S$) scale linearly with a factor $k$ by linearity of expectation:

\[
\metric(\mathcal{D, \text{I}}) = \frac{\mathop{\mathbb{E}}\limits_{0< j \le |\mathcal{B}|}k.S(j, \text{I}, \mathcal{D})}{\max\limits_{0< j \le |\mathcal{B}|}k.S(j, \text{I}, \mathcal{D})}
\]

The linear invariance comes from the fact that scaling the performance measure $Y_{LM}(x|(\text{I} + e_i))$ by a constant factor will not affect the ratio, as both the numerator and denominator will be scaled equally. Thus, \metric remains invariant to linear scaling.

\subsection*{Proposition 3: Invariance to Bin Selection}

If $S(j, \text{I}, \mathcal{D}) \approx \max\limits_{j}S(j, \text{I}, \mathcal{D})$ for all $i$, i.e., performance of the model is invariant to the bin from which the in-context examples are chosen, $\metric \to 1$.

\textbf{Proof:} Since, there is an equal probability to sample examples from every bin, 
\[
\mathop{\mathbb{E}}\limits_{0< j \le |\mathcal{B}|}\left[S(j, \text{I}, \mathcal{D})\right] = \frac{\sum_{j=1}^{|\mathcal{B}|}S(j, \text{I}, \mathcal{D})}{|\mathcal{B}|}\]

If the model's performance is agnostic to the bin from which the ICE are picked, formally
\[
\forall j \in \{1,..., \lvert B \rvert\},  S(j, \text{I}, \mathcal{D}) \approx \max\limits_{i}S(j, \text{I}, \mathcal{D})\]
\[
\frac{S(j, \text{I}, \mathcal{D})}{\max\limits_{j}S(j, \text{I}, \mathcal{D})} \approx 1
\]
\[
\metric = \frac{\mathop{\mathbb{E}}\limits_{0< j \le |\mathcal{B}|}S(i, \text{I}, \mathcal{D})}{\max\limits_{0< j \le |\mathcal{B}|}S(j, \text{I}, \mathcal{D})} \approx \frac{|\mathcal{B}|}{|\mathcal{B}|} = 1
\]

\subsection*{Proposition 4: Strict Retrieval Task Scenario}

Let $b_{\text{max}} = \arg\max_{j} S(j, \text{I}, \mathcal{D})$ be the best bin. If, $\forall$ $j \neq b_{\text{max}}$, $S(j, \text{I}, \mathcal{D}) \ll S(b_{\text{max}}, \text{I}, \mathcal{D})$, representing a strictly \textit{retrieval} task where the budget needs to be spent on in-context example optimization. In this scenario: $\mathrm{\metric} \to 1 / \lvert \mathcal{B} \rvert$ and $\lim\limits_{{\lvert \mathcal{B} \rvert \to \infty}} \mathrm{\metric} = 0$.

\textbf{Proof:} In this case, let the maximum score be denoted by, 
\[
m = S(b_{\text{max}}, \text{I}, \mathcal{D})
\]
If for $j \neq b_{\text{max}}$, $\frac{S(j, \text{I}, \mathcal{D})}{m} \ll 1$.  Assuming that we have a finite number of bins, the metric in this case reduces to 
\[
\metric = \frac{\mathop{\mathbb{E}}\limits_{0< j \le |\mathcal{B}|}S(i, \text{I}, \mathcal{D})}{\max\limits_{0< j \le |\mathcal{B}|}S(j, \text{I}, \mathcal{D})} \approx   \frac{m}{m.\lvert \mathcal{B} \rvert} =  \frac{1}{\lvert \mathcal{B} \rvert}
\]
Now, as the number of bins approaches infinity (for instance, in a scenario where examples are segregated based on their absolute distance from the query rather than percentile), the granularity of the partitioning becomes infinitely fine. Therefore,
\[
\lim\limits_{{\lvert \mathcal{B} \rvert \to \infty}} \mathrm{\metric} = 0
\]

\subsubsection{Discussion on the \metric Metric}
\label{app:metric-discussion}

We can define a partial order on the set of instructions as $I \preceq I'$ (in words, $I$ is \emph{pareto-dominated} by another instruction $I'$) if $S(j, I) \leq S(j, I')$, $\forall j$.
One can prove (under simple assumptions) that if there exists an $I^{*}$ that is \emph{not} Pareto-dominated by any other $I'$, then it must be a maximizer of $\sum_{j} w_{j} S(j, I)$ for some non-negative weights $w_{j}$'s \citep{harsanyi1955cardinal}. The weights $w_{j}$'s are typically uniform or chosen by domain experts \citep{harsanyi1955cardinal,gale1960theory}. For simplicity, we used  uniform weights to define our metric.

The bin-wise scores can also be used for instruction/prompt optimization. One possible way is as follows: Start with an initial instruction that maximizes $\sum_{j} S(j, I)$. Then, using weights $w_{j} \propto S(j, I)$, optimize $\sum_{j} w_{j} S(j, I)$ to find the next instruction, and iterate. This is almost an alternating optimzation over instruction and examples (if not examples, the weights over different bins from which examples are chosen) that intuitively optimizes instructions for the examples (or bins) that matter more for the task at hand. An optimized instruction should \textit{pareto-dominate} all other possible instructions. We leave this exploration and other uses of \metric for future work.

\subsection{Examples for the MTOP Dataset}

We show some examples from the MTOP dataset in Fig. \ref{fig:angelika}.

\begin{figure}[ht]
\centering
\begin{promptbox}

\texttt{Has Angelika Kratzer video messaged me?: \$[IN:GET\_MESSAGE [SL:CONTACT Angelika Kratzer ] [SL:TYPE\_CONTENT video ] [SL:RECIPIENT me ] ]\$} \\

\texttt{When will my next alarm start?: \$[IN:GET\_ALARM [SL:ORDINAL next ] ]\$} \\

\texttt{Ich möchte gerne mit meinen Cousinen Ashlyn, Linda, Chloe, und Angel über Whatsapp telefonieren: \$[IN:CREATE\_CALL [SL:CONTACT [IN:GET\_CONTACT [SL:CONTACT\_RELATED meinen ] [SL:TYPE\_RELATION Cousinen ] [SL:CONTACT Ashlyn ] [SL:CONTACT Linda ] [SL:CONTACT Chloe ] [SL:CONTACT Angel ] ] ] [SL:NAME\_APP Whatsapp ] ]\$} \\
\end{promptbox}
\caption{Some examples from the MTOP dataset.}
\label{fig:angelika}
\end{figure}

\subsection{\metric Results for Other Models}
\label{app:open-source-results}

\begin{figure*}[!t]
    \centering
    \includegraphics[trim={0cm 0cm 0cm 1cm},clip,width=\linewidth]{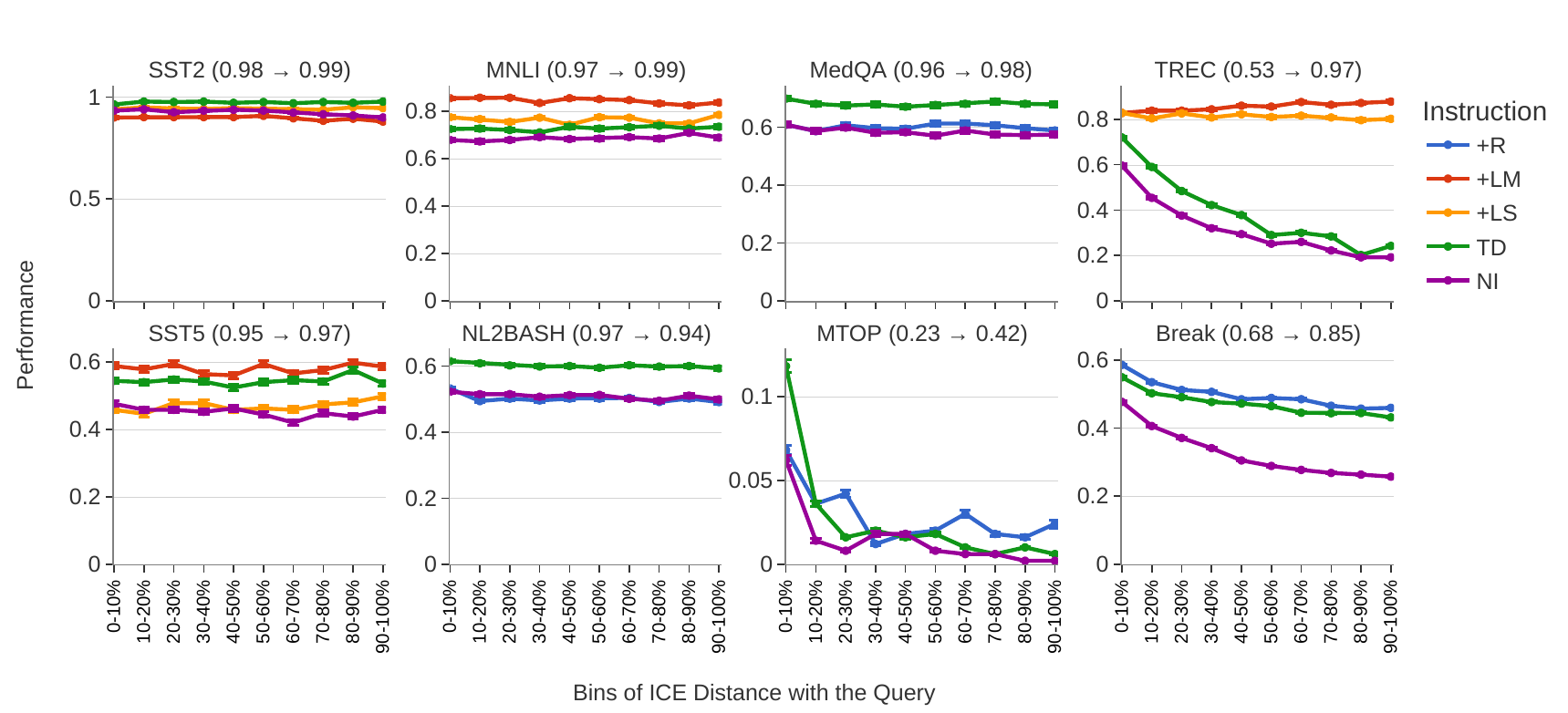}
    \caption{Per-bin performance trend of various datasets with distance-based bin using \texttt{GPT-3.5-Turbo}. Arrow indicates change in \metric from no instruction to a detailed instruction.}
    \label{fig:expt-1-gpt35}
\end{figure*}

\begin{figure*}[t]
    \centering
    \includegraphics[trim={0cm 0cm 0cm 1cm},clip,width=\linewidth]{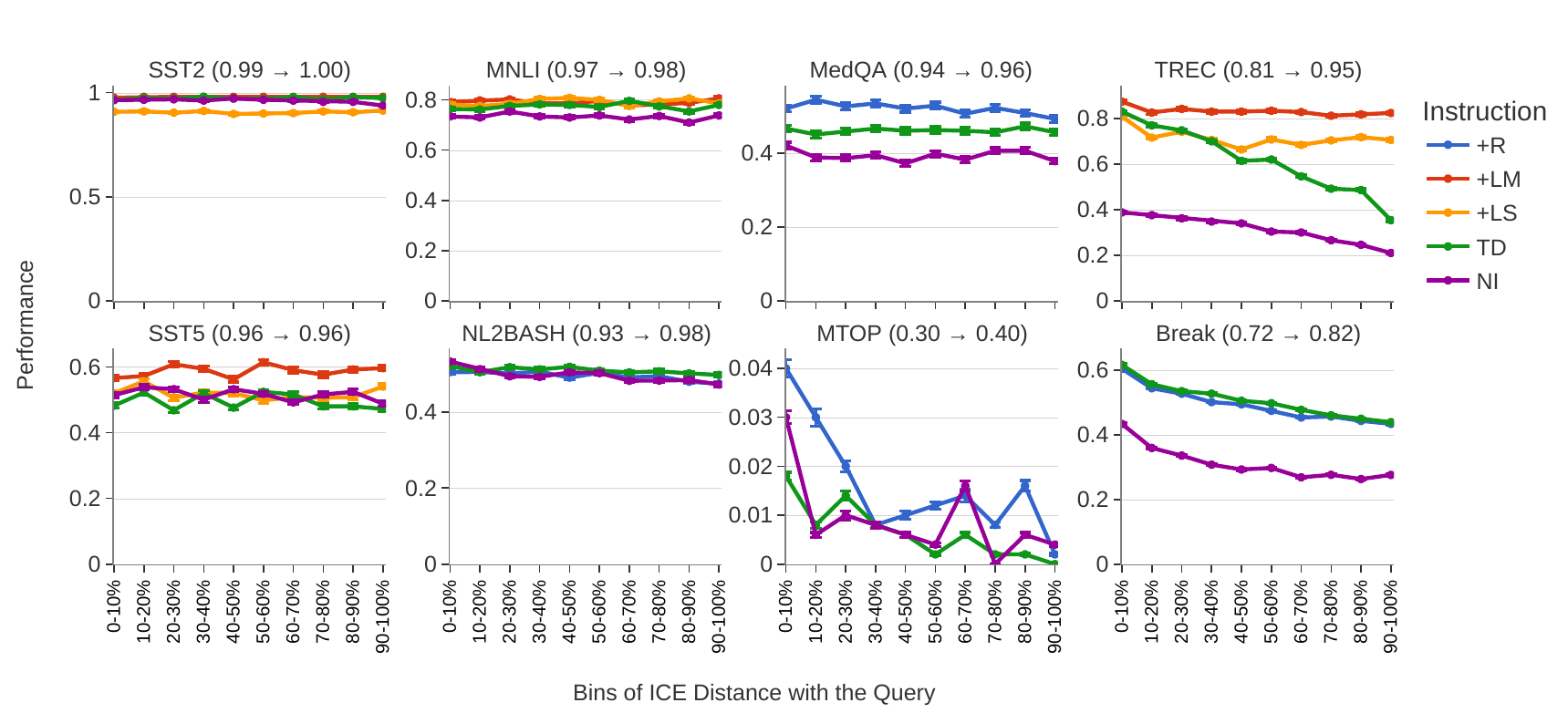}
    \caption{Per-bin performance trend of various datasets with distance-based bin using \texttt{Mixtral-8x7B-v0.1}, a mixture-of-experts model. Arrow indicates change in \metric from no instruction to a detailed instruction.}
    \label{fig:expt-1-mixtral}
\end{figure*}

\begin{figure*}[ht]
    \centering
    \includegraphics[trim={0cm 0cm 0cm 1cm},clip,width=\linewidth]{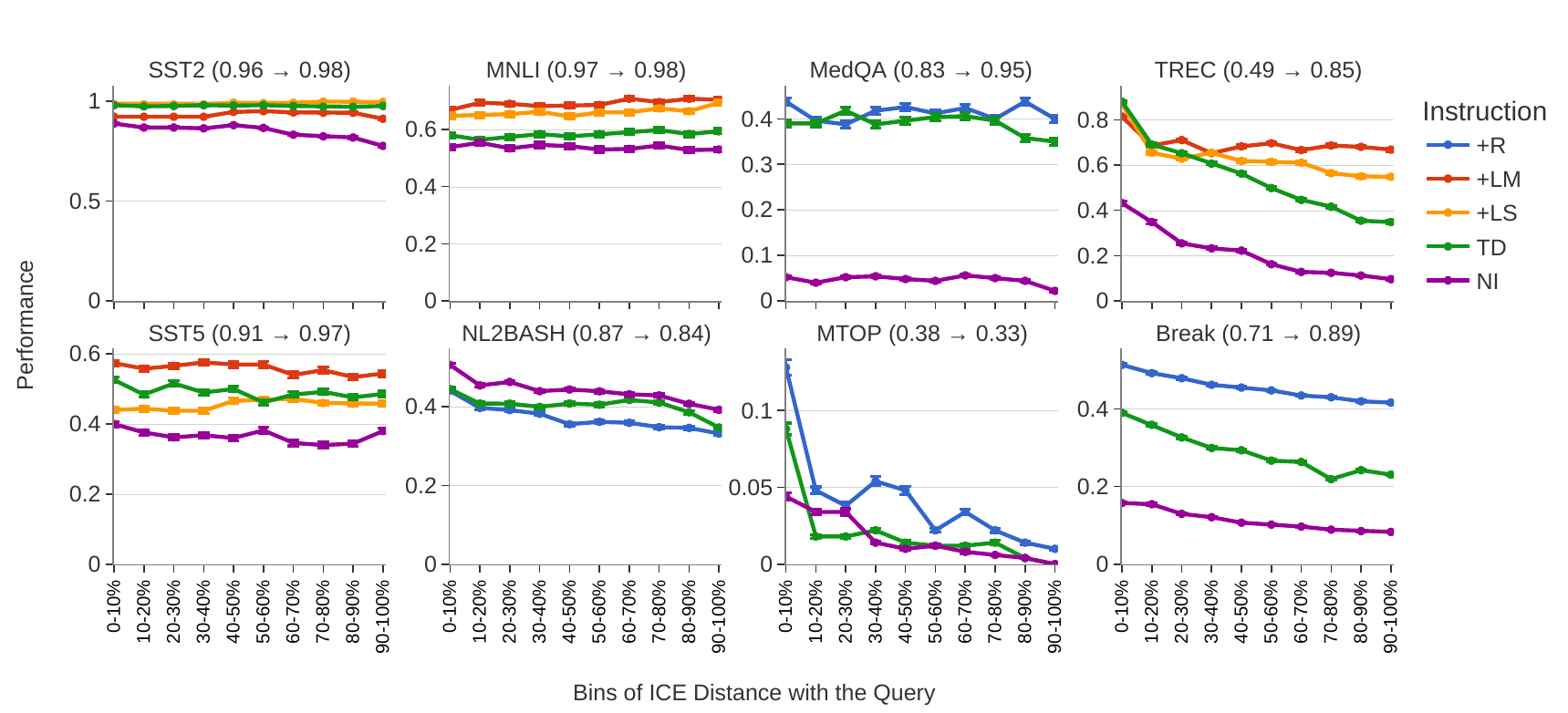}
    \caption{Per-bin performance trend of various datasets with distance-based bin using \texttt{Llama-2-70b-chat}. Arrow indicates change in \metric from no instruction to a detailed instruction.}
    \label{fig:expt-1-llama2}
\end{figure*}

\metric-metric results for various datasets with and without instructions for \texttt{GPT-3.5-Turbo} are shown in Fig. \ref{fig:expt-1-gpt35} and for \texttt{Mixtral-8x7B-v0.1} (an 8x7B mixture-of-experts model) and for \texttt{Llama-2-70b-chat} are shown in Figs. \ref{fig:expt-1-mixtral} and \ref{fig:expt-1-llama2}. We note similar observations as \texttt{GPT-4-Turbo} when using open-source language models.

\subsection{Pre-trained Models: Is Instruction-tuning the key?}

\begin{figure*}[!t]
    \centering
    \includegraphics[width=0.7\textwidth]{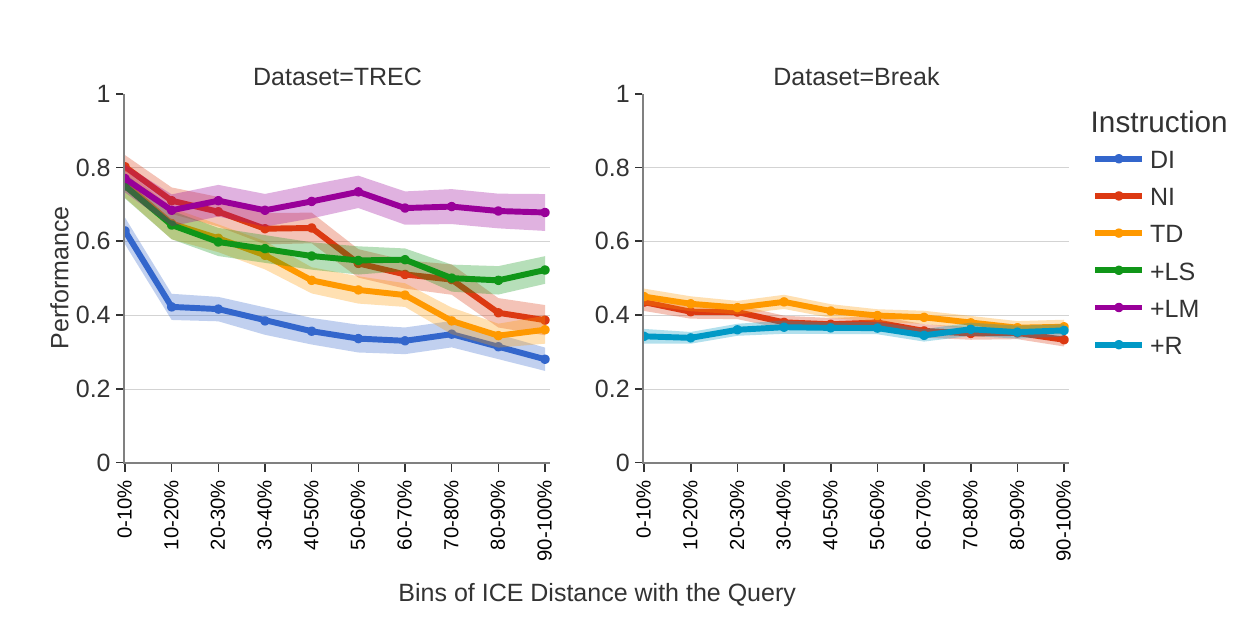}
    \caption{Comparing different instructions on TREC and Break using pre-trained \texttt{Llama-2-13B}. Abbreviation key same as in Fig. \ref{fig:lca}.}
    \label{fig:pretrained}
\end{figure*}

While the results demonstrated in Figs. \ref{fig:expt-1}, \ref{fig:expt-1-llama2}, and \ref{fig:expt-1-mixtral} validate our hypotheses for instruction-tuned models, we see in Fig. \ref{fig:pretrained} that they are valid for pre-trained models as well. Pre-training on next-token prediction enables models to learn a probability distribution $\text{p}_{\text{LM}}(\mathbf{x})$ over variable length sequences, thus allowing them to learn task-specific priors empowering them with zero-shot learnability, as was observed by \cite{choi2022lmpriors}. We see that even models not explicitly fine-tuned on instruction-following exhibit the ability to learn from contexts. 

We hypothesize that incorporating \textit{task-specific} information containing label space information and other contextual cues provide substantial improvement for LLMs to predict the next tokens more accurately.

In Fig. \ref{fig:pretrained}, we show the results of the \texttt{Llama-2-13B} model on TREC (classification) and Break (generation) tasks. We see a partial pareto-dominance in case of 
semantic classification task like TREC (with high \metric) while it is not observed for the \textit{schema-based} tasks like Break which has a low \metric.

\subsection{Additional Ablations and Results}
\label{app:additional-results}

\subsubsection{Task Transformations}
\label{app:task-transformations}

For further exploration into the kinds of tasks where ICE optimization matters, we designed the following ablation experiment: for the tasks with a standard output format (such as SST), we performed a transformation to the output space to induce a novel and unseen schema/structure to the output. Specifically, we defined the following mapping for the labels:

\begin{itemize}
    \item For SST2, we map:
    \begin{itemize}
        \item negative $\rightarrow$ negative
        \item positive $\rightarrow$ POSITIVE
    \end{itemize}
    \item For SST5, we map:
    \begin{itemize}
        \item very negative $\rightarrow$ very negative
        \item negative $\rightarrow$ negative
        \item neutral $\rightarrow$ NeUtRaL
        \item positive $\rightarrow$ POSITIVE
        \item very positive $\rightarrow$ VERY POSITIVE
    \end{itemize}
\end{itemize}

Since the new output labels are non-standard, we hypothesize that it would help to include similar in-context examples in the prompt, which are likely to share the same label, thus ensuring that the model is aware of the new label format. The observations from this ablation further strengthen our hypothesis.

The \metric scores with and without instructions in each case are given in Tab. \ref{tab:induced-schema}, and the original scores can be found in Tab. \ref{tab:dataset-details}.

\begin{table}[!ht]
\begin{center}
\begin{tabular}{lcc}
\hline
\textbf{Dataset} & \textbf{NI} & \textbf{DI} \\
\hline
SST2 (original) & 0.99 & 1.00 \\
SST2 (induced schema) & 0.86 & 0.96 \\
SST5 (original) & 0.91 & 0.94 \\
SST5 (induced schema) & 0.83 & 0.93 \\
\hline
\end{tabular}
\end{center}
\caption{\metric scores with and without instructions in original vs. induced schema on SST2 and SST5. Note the dip in performance without any instruction as we move to an induced schema.}
\label{tab:induced-schema}
\end{table}

Quantitatively, we observe the following:

\begin{itemize}
    \item A dip in the NICE score (corresponding to the No-Instructions case) as compared to standard datasets, thus showing the importance of learning the schema for model performance.
    \item This dip (showing the dependence of ICE choice) recovers in presence of detailed instructions (since these tasks are easy enough to be learned from \textit{detailed task-specific} instructions).
\end{itemize}

\subsubsection{Correctness Metrics on GSM8k}
\label{app:gsm8k}
We experiment with the GSM8k \cite{cobbe2021gsm8k} task to study the role of ICE optimization for complex mathematical reasoning tasks. We make use of 4-shot examples in each case. An example output for this dataset looks like this: 

\texttt{Natalia sold 48/2 = <<48/2=24>>24 clips in May. Natalia sold 48+24 = <<48+24=72>>72 clips altogether in April and May. \\
\#\#\#\# 72} 

We define two metrics to evaluate model outputs:

\textbf{Final Answer Correctness (FAC)} We consider the last numerical value in the string response to compare against the ground-truth and do not penalize the model for the format. The \metric scores with respect to the FAC metric are high ($0.91$ with no instructions to $0.98$ with detailed instructions), which means the task becomes almost invariant to choice of examples.


\textbf{Reasoning Format Correctness (RFC):} On the output string of the model, we use regular expressions to check if the model adheres to the reasoning syntax. In this case, we see that the \metric scores are $0.72$ (for no instructions) and $0.74$ (for detailed instructions), i.e., significantly lower than $0.91$/$0.98$, indicating that ICE optimization matters a lot for learning the correct format. This is consistent with our hypothesis about schema being the most important thing that is learned through examples.


\end{document}